\documentclass[lettersize,journal]{IEEEtran}
\usepackage{amsmath,amsfonts}
\usepackage{algorithmic}
\usepackage{algorithm}
\usepackage{array}
\usepackage[caption=false,font=normalsize,labelfont=sf,textfont=sf]{subfig}
\usepackage{textcomp}
\usepackage{stfloats}
\usepackage{url}
\usepackage{verbatim}
\usepackage{graphicx}
\usepackage{cite}
\usepackage{booktabs}
\usepackage{multirow}
\usepackage{multicol}
\usepackage{xcolor}

\begin{document}

\title{DP-IQA: Utilizing Diffusion Prior for Blind Image Quality Assessment in the Wild}

\author{
Honghao Fu\textsuperscript{\rm *}, Yufei Wang\textsuperscript{\rm *}, Wenhan Yang, \textit{Member, IEEE}, Alex C. Kot, \textit{Life Fellow, IEEE}, Bihan Wen\textsuperscript{\textdagger}, \textit{Senior Member, IEEE}
\thanks{Honghao Fu, Yufei Wang, Alex C. Kot and Bihan Wen are with the School of Electrical and Electronic Engineering, Nanyang Technological University, Singapre. (e-mail: \{hfu006, yufei001, eackot, bihan.wen\}@ntu.edu.sg)}
\thanks{Wenhan Yang are with the PengCheng Laboratory, China. (e-mail: yangwh@pcl.ac.cn)}
\thanks{\textsuperscript{\rm *}These authors contributed equally.}
\thanks{\textsuperscript{\textdagger}The corresponding author.}\\
}

\markboth{Journal of \LaTeX\ Class Files,~Vol.~14, No.~8, August~2021}%
{Shell \MakeLowercase{\textit{et al.}}: A Sample Article Using IEEEtran.cls for IEEE Journals}


\maketitle

\begin{abstract}
Blind image quality assessment (IQA) in the wild, which assesses the quality of images with complex authentic distortions and no reference images, presents significant challenges. Given the difficulty in collecting large-scale training data, leveraging limited data to develop a model with strong generalization remains an open problem. Motivated by the robust image perception capabilities of pre-trained text-to-image (T2I) diffusion models, we propose a novel IQA method, diffusion priors-based IQA (DP-IQA), to utilize the T2I model's prior for improved performance and generalization ability. 
Specifically, we utilize pre-trained Stable Diffusion as the backbone, extracting multi-level features from the denoising U-Net guided by prompt embeddings through a tunable text adapter. Simultaneously, an image adapter compensates for information loss introduced by the lossy pre-trained encoder.
Unlike T2I models that require full image distribution modeling, our approach targets image quality assessment, which inherently requires fewer parameters. To improve applicability, we distill the knowledge into a lightweight CNN-based student model, significantly reducing parameters while maintaining or even enhancing generalization performance.
Experimental results demonstrate that DP-IQA achieves state-of-the-art performance on various in-the-wild datasets, highlighting the superior generalization capability of T2I priors in blind IQA tasks. To our knowledge, DP-IQA is the first method to apply pre-trained diffusion priors in blind IQA. The code is available at \url{https://github.com/RomGai/DP-IQA}.
\end{abstract}

\begin{IEEEkeywords}
Blind IQA, diffusion prior, text-to-image model, knowledge distillation.
\end{IEEEkeywords}

\section{Introduction}
\label{sec:intro}
\IEEEPARstart{M}{illions} of images are uploaded and spread across the internet daily~\cite{Intro1}. Inevitably, some of these images are of poor quality, causing negative impressions due to their visual defects~\cite{Intro2}. Image Quality Assessment (IQA) evaluates the visual quality of images from a human perspective, to ensure high-quality content for applications such as social media sharing and streaming~\cite{ReIQA}. Therefore, the robustness and generalization of IQA methods against various real-world distortions significantly impact the presentation of billions of images to the public.
Blind IQA (BIQA) methods, also known as no-reference IQA, are crucial for evaluating image quality without reference images. In diverse and uncontrolled real-world environments \textbf{(``in-the-wild'')}, BIQA is particularly necessary due to the unpredictable distortions present. Unlike methods that require reference images, BIQA directly predicts image quality, which is essential for handling authentic distortions. However, labeling the dataset to train BIQA models is laborious because it requires multiple volunteers to provide subjective scores for each image to avoid bias, resulting in a smaller scale of the dataset compared to other tasks like image classification~\cite{KonIQ,NNID}.

\begin{figure}[t]
  \centering
  \includegraphics[width=0.70\linewidth]{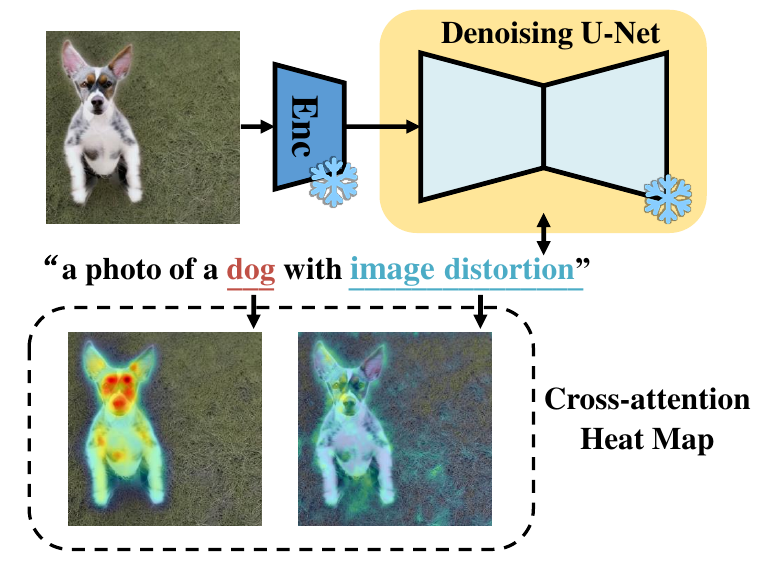}
  \caption{The motivation of our work. Unlike commonly used classification priors for IQA, which only rely on image and category labels and focus on high-level, instance-level features, T2I models benefit from extensive, diverse training data that includes both high- and low-quality images with corresponding text prompts. As shown in the figure above, this enables T2I models to capture both high-level semantic features and low-level distortions simultaneously, making them a more effective prior for blind IQA.}
  \label{moti}
\end{figure}


To increase the generalization ability of BIQA models under limited data, the majority of recent BIQA methods~\cite{MUSIQ,TReS, DEIQT,ReIQA,LoDa,ARNIQA} leverage priors from pre-trained image classification models. These priors emphasize high-level vision and consequently lack adequate low-level information, which creates potential barriers and increases the difficulty for the model in learning low-level features. This issue arises because, during classification training, images with similar high-level content but differing low-level quality are assigned the same label~\cite{QPT}. Furthermore, using networks pre-trained for classification does not align well with human visual perception of image quality~\cite{LIQE}. Humans can recognize and classify objects in an image even if it is distorted, as long as the distortion is not too severe. Therefore, recent research~\cite{CLIPIQA, LIQE, IPCE} leverages the prior knowledge of visual-language multimodal models for BIQA tasks, reducing reliance on classification priors. An advanced approach involves constructing a set of text templates that describe both the high-level content and low-level quality of the input images, and utilizing the visual-language model CLIP~\cite{CLIP} to obtain feature embeddings for both the image and corresponding text. The similarities among them are used as metrics to further measure the image quality.
%
However, recent research reveals that CLIP image encoder is largely insensitive to various distortion types~\cite{DACLIP}, demonstrating effective performance only with a limited set of distortions (blurry, hazy, and rainy). In contrast, its text encoder proficiently manages related textual descriptions. This discrepancy results in a mismatch between the CLIP embeddings of distorted images and the clean descriptions of distortion types~\cite{DACLIP}. Furthermore, the CLIP image encoder compresses complex images into vectors, potentially leading to the loss of low-level information. Therefore, the current methods utilizing CLIP priors for BIQA still have limitations. This prompts us to explore whether BIQA could benefit from more ideal priors offered by other tasks and models.

As shown in Figure~\ref{moti}, inspired by the robust image perception capabilities of text-to-image (T2I) diffusion models, we propose leveraging diffusion priors for blind IQA (BIQA). While a few recent studies have explored using diffusion models~\cite{PFD-IQA,DiffV^2IQA} for BIQA, they still rely on pre-trained classification models and do not fully utilize the large-scale pre-trained T2I priors. Priors from pre-trained T2I diffusion models have been effectively applied to high-level tasks such as image classification~\cite{SDYour} and semantic segmentation~\cite{DiffSeg,VPD}, as well as low-level tasks like super-resolution~\cite{StableSR} and image restoration~\cite{GenerativePrior,ShadowDiffusion}. This further confirms that diffusion priors encompass a rich blend of high-level and low-level information.
Furthermore, employing a T2I model like Stable Diffusion (SD)~\cite{SD} avoids processing distorted images through the CLIP image encoder, which is insensitive to various distortions. Instead, it only utilizes the CLIP text encoder to condition the T2I model, which can accurately embed text descriptions of image distortions. 
Additionally, incorporating negative prompts during inference with T2I models to prevent the generation of undesirable image content is a widely adopted engineering practice. These negative prompts typically include descriptions related to image quality, such as ``blurry'', ``low resolution'', ``worst/low/normal quality'', and ``JPEG artifacts.'' This suggests that T2I models are capable of recognizing the components associated with these quality-related and distortion-related prompts in the images. 
%
However, despite these advantages, unlike IQA methods based on pre-trained classification models or CLIP, which can directly obtain feature vectors from the models' output layer, how to effectively extract features for IQA tasks from T2I diffusion models remains an open problem.

In this paper, we explore the potential of T2I diffusion models and adapt them to better address in-the-wild BIQA with various unpredictable authentic distortions. We propose a novel BIQA method called diffusion prior-based IQA (DP-IQA). DP-IQA leverages a pre-trained SD model as the backbone, extracting multi-level features from the denoising U-Net at a specific timestep and decoding them to estimate image quality, without requiring a whole diffusion process. {From a practical perspective, a text adapter is used to address the potential domain gap caused by our constant conditional embedding strategy, while an image adapter~\cite{T2IADP} supplements features from the original image to bypass the distortion information bottleneck of the variational autoencoder (VAE).} To more effectively utilize the T2I model’s image understanding and global modeling capabilities, DP-IQA processes the entire image without patch splitting, allowing for better extraction of semantic features. Unlike T2I models that require full image distribution modeling, our approach focuses on image quality assessment, which inherently requires fewer parameters. {Consequently, we distill the knowledge from this model into a CNN-based student model, significantly reducing parameters to enhance its practicality in real-world applications.} Experiments demonstrate that DP-IQA achieves state-of-the-art (SOTA) performance and superior generalization ability across various in-the-wild datasets. To the best of our knowledge, DP-IQA is the first method to apply T2I diffusion priors in BIQA. Our contributions are summarized as follows:

\begin{itemize}
    \item We are the first to leverage the pretrained T2I diffusion model's prior for blind IQA, specifically its strong ability to model semantic and low-level features simultaneously.
    \item We propose a framework that can better extract aesthetics-related features from activation values during the diffusion denoising step, resulting in a more compact and effective representation for subsequent prediction. Besides, the enhanced T2I diffusion priors are distilled into a lightweight model for enhanced applicability, achieving $\sim3\times$ speed up and $\sim14\times$ reduction in parameters under similar performance.
    \item The extensive experiments demonstrate the effectiveness and generalization ability of the proposed method on several in-the-wild benchmarks with authentic distortions.
\end{itemize}

\section{Related Works}
\label{sec:rw}

\subsection{Blind image quality assessment}

Traditional BIQA primarily leverages statistical features from the spatial and transform domains of images using natural scene statistics\cite{2001, 2002, 2004} and employs machine learning models for the regression of image quality score\cite{201, 202, 203, MTR}. However, these methods often fail to capture high-level image information due to their reliance on specific feature computations. Recently, deep learning has advanced BIQA significantly~\cite{205, 2011, 2015, DBCNN, PeIQA, MetaIQA, QPT}. Initial methods used Convolutional Neural Networks (CNNs)~\cite{fu2023sgcn} to learn image quality features~\cite{207, 208}, while recent works~\cite{TIQA, MUSIQ} propose to leverage powerful Vision Transformer (ViT)~\cite{ViT} for better performance.

To address the challenge posed by the limited scale of IQA datasets hindering the models' representational capabilities, utilizing priors from classification models pre-trained on larger-scale image datasets like ImageNet is a commom practice~\cite{2013,209,210,211,GraphIQA,HyperIQA,TReS,TTL,DEIQT, LoDa,QPT,ARNIQA,CDI,NROU}. However, as discussed in the previous section, it exhibits significant differences from human visual perception habits. There are also some works avoid using pre-trained classification models. For example, early generative models such as Generative Adversarial Networks (GANs) have been applied to IQA tasks~\cite{212, 213, 214}. GAN-based methods typically reconstruct an undistorted image from a distorted one, then extract features from this process, or use the reconstructed image as a reference for IQA. Consequently, they require undistorted reference images during training, which limits their applicability to in-the-wild images without references. More recent works, such as CLIP-IQA~\cite{CLIPIQA}, LIQE~\cite{LIQE} and IPCE~\cite{IPCE}, adopt the priors of vision-language model CLIP for BIQA~\cite{LGP}. They perform IQA by minimizing the cosine similarity between the CLIP embedding of the image and the CLIP embedding of text describing its content and quality. However, as stated in the previous section, the CLIP image encoder is not sensitive to a large number of distortion types, while its text encoder can accurately embed text describing these distortions, leading to a mismatch between image and text embeddings~\cite{DACLIP}. Therefore, applying CLIP priors to in-the-wild BIQA may still have limitations.

Recently, a few studies have applied diffusion models to BIQA. PFD-IQA~\cite{PFD-IQA} trains a diffusion model to denoise prior features of images obtained through pre-trained ViT and performed regression on the denoised features to predict quality scores. {GenzIQA~\cite{GenzIQA} utilizes cross-attention maps from a pretrained diffusion model as quality representations and employs prompt tuning to adapt the conditional embeddings for the IQA task.} Diff$V^2$IQA~\cite{DiffV^2IQA} trains a diffusion model on 2 small-scale synthetic distortion datasets to restore distorted images to high-quality images, and uses ViT and ResNet~\cite{RN50} to obtain the features of intermediate denoised images from the denoising process to predict quality scores. However, due to the poor performance of its self-trained diffusion model, the restored images significantly deviate from the original images, introducing new distortions not accounted for in the datasets’ scoring system. Additionally, since the synthetic distortion datasets contains only a limited number of distortion types, the self-trained diffusion model lacks robustness to complex real-world distortions.


\begin{figure*}[t]
  \centering
  \includegraphics[width=0.88\linewidth]{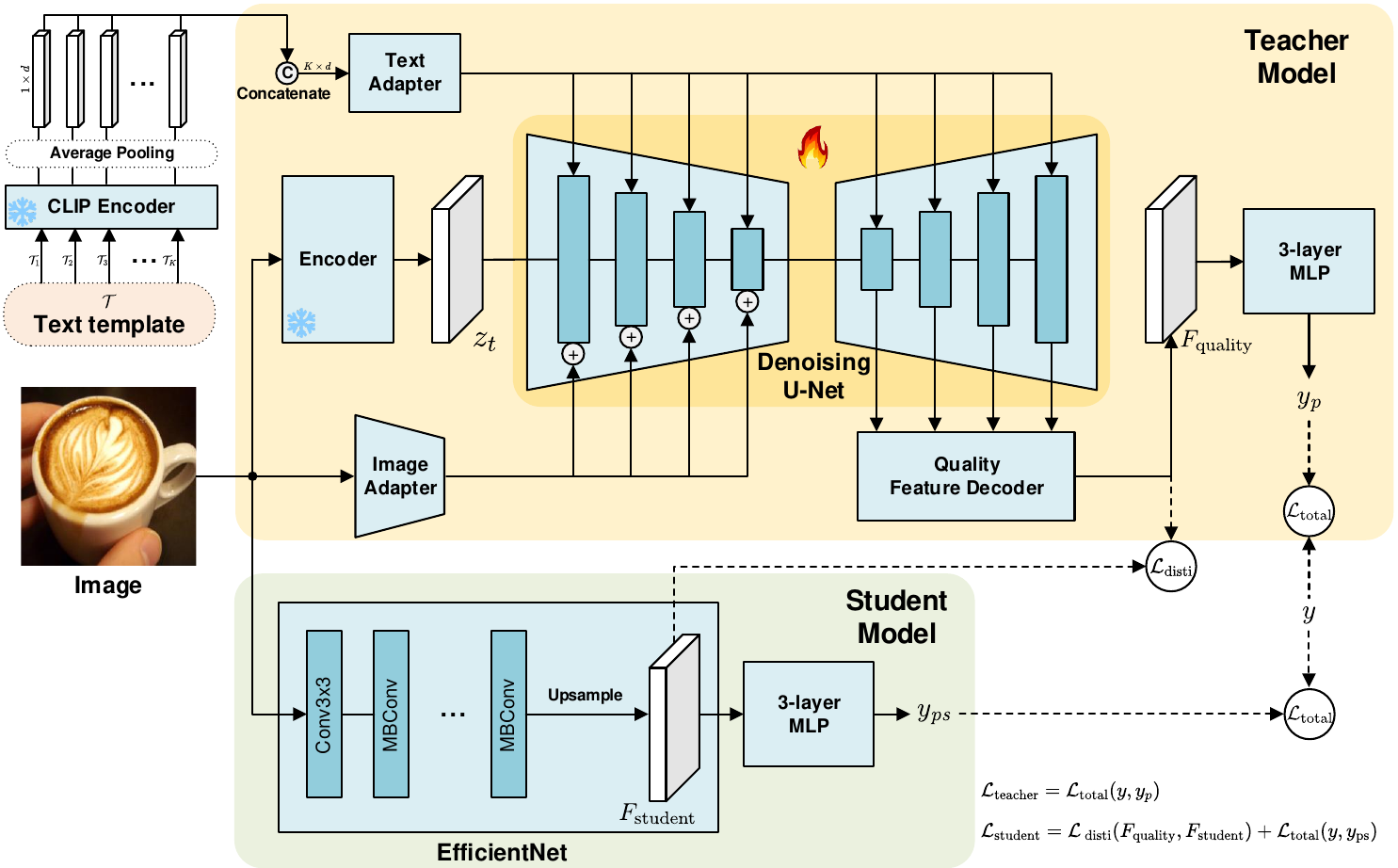}
  \caption{
  Framework of DP-IQA and its corresponding student model by knowledge distillation. 
  The pretrained CLIP encoder is first used to convert quality assessment-related text templates into embeddings, serving as the condition for the denoising U-Net. A text adapter is employed to bridge the gap introduced by fixed text templates, while an image adapter is incorporated to preserve low-level features essential for IQA. Finally, a Quality Feature Decoder (QFD) is designed to fuse features across multiple levels, producing the final feature maps, which are then passed through an MLP to obtain the final result. Unlike diffusion models, we generate the feature maps in a single feedforward pass by using a fixed timestep.
  To further improve the efficiency, we distill the knowledge from DP-IQA into a student model with EfficientNet as the backbone to further reduce parameters and increase inference speed. 
  The loss functions are detailed in equation (\ref{l_teacher}) and (\ref{l_student}).}
  \label{fw}
  \vspace{-0.1in}
\end{figure*}

\subsection{Diffusion model priors}

Diffusion-based generative models excel in generating high-quality images with intricate scenes and semantics from textual descriptions, demonstrating a profound understanding of text and vision. The prior knowledge embedded in large-scale pre-trained diffusion models like SD has proven effective for high-level visual tasks such as image classification~\cite{SDYour}, semantic segmentation~\cite{DiffSeg, VPD}, and depth estimation~\cite{VPD, Marigold}. Additionally, it has also been utilized in low-level tasks like super-resolution~\cite{StableSR} and image restoration~\cite{DreamClean, GenerativePrior,zhang2024defending,zhang2024tuning,gu2023orsi,xu2024cascade,fu2023brainvis, ShadowDiffusion}, showing impressive results. This indicates that the diffusion priors contain sufficient high-level and low-level information with no significant barriers between them. However, the challenge of fully harnessing its ability to represent visual features for image quality assessment and effectively utilizing its prior knowledge remains an open problem. Recent studies have provided compelling evidence~\cite{l-DAE} suggesting that the representational capability of  diffusion models is predominantly derived from the denoising-driven process rather than the diffusion-driven process. Consequently, we focus on exploring strategies to effectively leverage prior knowledge embedded in the denoiser within the diffusion pipeline.


\section{Methodology}
\label{sec:method}
\subsection{Preliminary}
\textbf{Diffusion.} As the backbone of our proposed DP-IQA, we first provide a brief introduction to the principles of diffusion models. Let $z_t$ be the random noise at the $t$-th timestep. Diffusion models transform $z_t$ to the denoised sample $z_0$ by gradually denoising $z_t$ to a less noisy $z_\text{t-1}$. The forward diffusion process is modeled as:
\begin{equation}
q(z_t \mid z_{t-1}) = \mathcal{N}(z_t; \sqrt{\alpha_t} \, z_{t-1}, (1 - \alpha_t)\mathbf{I}),
\end{equation}
where $\{\alpha_t\}$ are fixed coefficients that determine the noise schedule. {By defining} $\bar{\alpha}_t = \prod_{s=1}^t \alpha_s$, $z_t$ can be obtained directly from $z_0$~\cite{Label-efficient}:
\begin{align}
    q(z_t \mid z_0) = \mathcal{N}(z_t; \sqrt{\bar{\alpha}_t} \, z_{t-1}, (1 - \bar{\alpha}_t)\mathbf{I}),\\
    z_t = \sqrt{\bar{\alpha}_t} \, z_0 + \sqrt{1 - \bar{\alpha}_t} \, \epsilon, \quad \epsilon \sim \mathcal{N}(0, \mathbf{I}). 
\end{align}
It makes sampling for any $z_t$ more efficient. With proper re-parameterization, the training objective of diffusion models can be derived as~\cite{DM,VPD}:
\begin{equation}
\mathcal{L}_{\text{DM}} = \mathbb{E}_{z_0, \epsilon, t} \left[ \| \epsilon - \epsilon_\theta (z_t(z_0, \epsilon), t; \mathcal{C}) \|_2^2 \right],
\end{equation}
where $\epsilon_\theta$ is a denoising autoencoder that is learned to predict $\epsilon$ given the conditional embedding $\mathcal{C}$. In our task, the denoising autoencoder $\epsilon_\theta$ is a U-Net, $z_t$ is a latent representation of a distorted image, which can also be regarded as a latent variable that has not been fully denoised from random noise. By controlling the conditional embedding $\mathcal{C}$, we enable the denoising U-Net to effectively extract different features from $z_t$, and thereby extract the prior knowledge required for the IQA task from a single timestep in the diffusion process.

\subsection{Overview}
\begin{figure}[t]
  \centering
  \includegraphics[width=\linewidth]{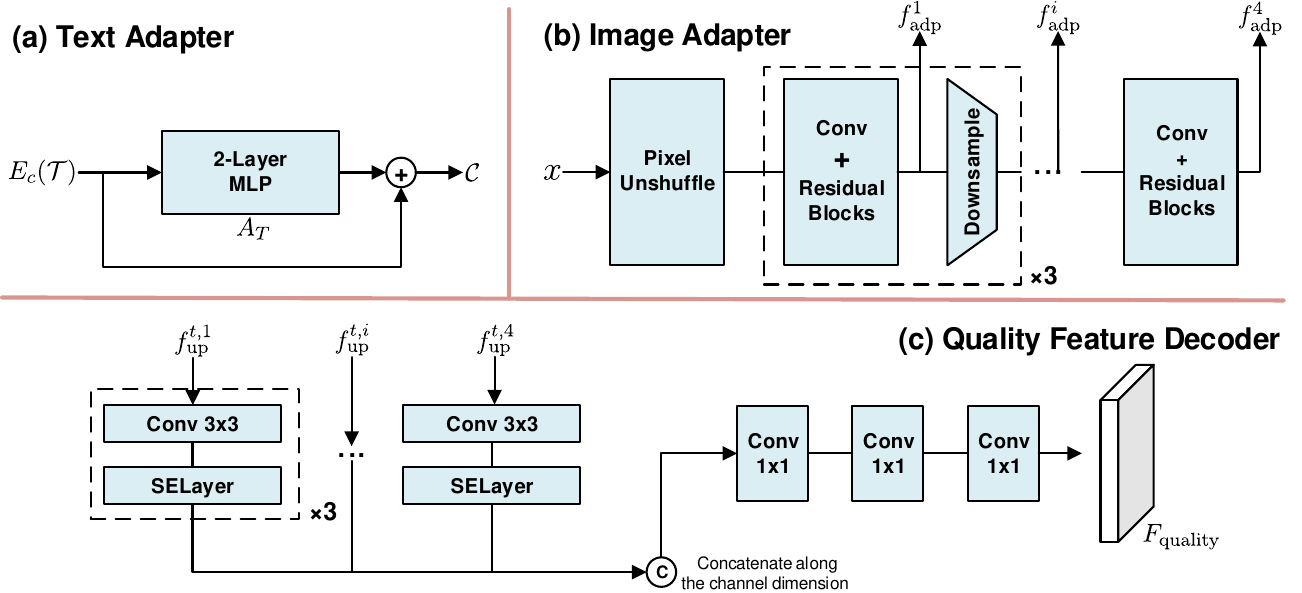}
  \caption{Details of the (a) text adapter, (b) image adapter~\cite{3-4} and (c) quality feature decoder in DP-IQA. {The image adapter is a lightweight module designed to complement the potential loss of degradation-specific information in the pretrained encoder of Stable Diffusion.}}
  \label{dt}
\end{figure}

We adapt the representation capabilities and priors of T2I diffusion models to BIQA \textbf{in the wild}, as illustrated in Figure~\ref{fw}.
Specifically, the input image is first encoded with a pre-trained VAE encoder, then fed into the denoising U-Net of the pre-trained SD~\cite{SD}. Concurrently, a CLIP encoder~\cite{CLIP} converts text describing the image quality into conditional embeddings for the denoising U-Net. The input text is templated and consistent across all images. {Meanwhile, text and image adapters are adopted to mitigate the domain gap caused by the constant conditional embedding strategy} and correct the information loss caused by the VAE bottleneck. Subsequently, we extract feature maps from each stage of the U-Net's upsampling process, which are then fused and decoded by a well-designed Quality Feature Decoder (QFD). Finally, a Multi-Layer Perceptron (MLP) is employed to regress the image quality scores. Figure~\ref{dt} provides details on the adapters and QFD. 
After obtaining the above teacher model, we distill the knowledge in the trained DP-IQA into an EfficientNet-based~\cite{EN} student model, which is initialized with the official pre-trained weights, and its output structure is modified to align with the teacher model. The distillation process leverages two sources of supervision: (1) the output feature map from the QFD, and (2) the GT image quality scores.

\label{teacher}
\noindent\textbf{Extracting diffusion priors from a single timestep.} A pre-trained T2I diffusion model contains sufficient information to sample from the data distribution, including its low-level features and structures, as the model can be viewed as the learned gradient of data density~\cite{VPD}. With limited natural language supervision during pre-training, the T2I model also incorporates significant high-level knowledge. {Recent work l-DAE~\cite{l-DAE} shows that the representational power of denoising diffusion models primarily stems from the denoising process rather than the diffusion itself. This suggests that a single-step denoising is sufficient to leverage the representation capability of the pretrained denoiser, without requiring a diffusion process. Thus, we adopt a single-timestep setting for our task.} We utilize the pre-trained SD as our backbone. Assume we wish to utilize the diffusion priors expressed by the denoising U-Net $\epsilon_\theta$ at timestep $t$. For an input image $x \in \mathbb{R}^{H \times W \times 3}$, it is encoded into latent representation $z_t$ by a pretrained VAE. Then, from $\epsilon_\theta(z_t, t)$, we obtain the feature maps $f_{\text{up}}^i$ at each upsampling stage, where $i=1,2,3,4$. The resulting set of feature maps $F_{\text{up}}^t = \{f_{\text{up}}^{t,1}, f_{\text{up}}^{t,2}, f_{\text{up}}^{t,3}, f_{\text{up}}^{t,4}\}$ is the prior features at $t$. {Such a multi-level feature extraction strategy facilitates a more comprehensive representation of the image, as the upsampling process of the denoising U-Net transitions features from low-resolution, high-level semantics to high-resolution, fine-grained details. Moreover, since multi-level features from the downsampling stages are fused into the upsampling path via skip connections, this strategy also effectively captures information propagated during the downsampling process.}

\noindent\textbf{Text template.} In a T2I diffusion model, text is converted into conditional embeddings by a text encoder to guide the denoising process. SD uses a CLIP encoder for embedding text. An appropriate text prompt is crucial for the denoiser to focus on the target features. We use a general text template summarized by previous MLLM-based arts~\cite{LIQE,QBench,QBenchpp} to describe the image's content and quality as the text conditional input. The template is ``a photo of a \{scenes\} with \{distortion type\} distortion, which is of \{quality level\} quality.'' \{Scenes\} includes typical image subjects, \{distortion type\} covers common distortions, and \{quality level\} provides a general quality description, like ``bad'' or ``good''. Both scenes and distortion types also offer general descriptions. For example, ``a photo of an animal with realistic blur distortion, which is of bad quality.'' {However, real-world distortions are often complex and cannot be fully captured by a limited set of labels. To ensure robustness, we use the category ``other'' to represent distortions not explicitly defined in our templates.} Assuming there are $l_s$ scenes, $l_d$  distortion types, and  $l_q$ quality levels, there are a total of $K = l_s \cdot l_d \cdot l_q$ combinations. We define $\mathcal{T}$ as the set of all combinations, where $\mathcal{T}_k$ is the $k$-th sentence in $\mathcal{T}$.

\noindent{\textbf{Constant conditional embedding.} 
{In T2I models, the text is typically split into a sequence of tokens, each of which is encoded into a conditional embedding. These embeddings are then used in the U-Net's cross-attention mechanism to guide the model's focus toward the content described in the text and its corresponding image features.} Benefiting from this, our method does not require setting specific text template content for each input image. Instead, it inputs all the template combinations simultaneously. In the CLIP encoder $E_C$ of the T2I diffusion model, an input prompt is split into multiple tokens (77 in SD by default, which can be modified). Define the output dimension of $E_C$ as $d$, then each token is converted into an $1 \times d$ embedding. The embeddings of all tokens are concatenated ($77 \times d$) as the condition, influencing the attention mechanism. This allows us to treat each sentence in our template as a separate token, combining them into a universal constant condition embedding in our task, prompting the U-Net to be able to focus on all the distortion scenarios it needs to pay attention to. In practice, each sentence is first split into tokens, and the embeddings of all its tokens are average pooled to produce a vector with the same shape as the embedding of a single token ($1 \times d$), which represents the global embedding of a sentence. The pooled results of $K$ sentences from the template $\mathcal{T}$ are then concatenated to form an overall embedding with shape $K \times d$, which is simplified to $E_C(\mathcal{T}) \in \mathbb{R}^{K \times d}$. $E_C(\mathcal{T})$ will be used as a constant in our task to provide a universal conditional embedding.
Therefore, we can apply all combinations of text template to an image at once, which helps the model better understand an image with multiple scenes and distortions}

\subsection{Diffusion prior-based IQA (DP-IQA)}

\noindent\textbf{Text adapter.} However, our text template and conditional embeddings slightly differs from the standard strategy of pre-trained SD, which may lead to the potential domain gap. {Previous work~\cite{clipadapter} has shown that compared to learning task-specific prompts, introducing adapters can improve CLIP's performance on new tasks beyond the pre-training setting while also reducing computational costs.} Therefore, we use a text adapter~\cite{VPD, 3-2, 3-3} to mitigate this gap. CLIP-Adapter~\cite{clipadapter}, which systematically demonstrates that compared to learning task-specific prompts, introducing adapters can improve CLIP's performance on new tasks beyond the pre-training setting while also reducing computational costs.
The text adapter consists of a two-layer MLP $A_T$, and takes $E_C(\mathcal{T})$ as input. The output of $A_T$ is then added to $E_C(\mathcal{T})$ to obtain the adjusted conditional embedding $\mathcal{C} \in \mathbb{R}^{K \times d}$. This process is: 
\begin{equation}
\mathcal{C} = E_C(\mathcal{T}) + A_T(E_C(\mathcal{T})).
\end{equation}

\noindent\textbf{Image adapter.} 
{Previous work has shown that VAEs act as a lossy compression method~\cite{vlae}, typically encoding only information that cannot be locally reconstructed into the latent space, such as long-range dependencies. Specifically, VAEs preserve only global structures in the latent variables, while local details are modeled directly by the decoder. This results in the loss of low-level details during the encoding process. In addition, some diffusion-based works focused on low-level vision have also found that VAEs are not robust to distortions under standard training conditions~\cite{SUPIR,DiffBir}. However, fine-tuning a VAE adapted for a pretrained diffusion pipeline on low-level tasks is prohibitively expensive. To address this, we introduce an image adapter $A_I$ that bypasses the VAE's compression process and directly extracts features from the original image $x$ as a supplement.} These features are fed into the denoising U-Net's downsampling process.
%
Define the feature map at each downsampling stage as $f_{\text{down}}^i$, where $i=1,2,3,4$. The set of the feature maps at timestep $t$ is $F_{\text{down}}^t = \{f_{\text{down}}^{t,1}, f_{\text{down}}^{t,2}, f_{\text{down}}^{t,3}, f_{\text{down}}^{t,4}\}$. Define the output of the image adapter as $A_I(x) = F_{\text{adp}}^i = \{f_{\text{adp}}^1, f_{\text{adp}}^2, f_{\text{adp}}^3, f_{\text{adp}}^4\}$, which is independent of the timestep $t$, and the size of $f_{\text{adp}}^i$ is consistent with $f_{\text{adp}}^{t,i}$. The process of feature supplementation by the image adapter is:
\begin{equation}
F_{\text{down}}^{t,i} = F_{\text{down}}^{t,i} + F_{\text{adp}}^i, \quad i = 1, 2, 3, 4.
\end{equation}

\noindent\textbf{Quality feature decoder (QFD).} We design a CNN-based QFD $D$ to decode the feature maps from the upsampling stages, and then regress the output of the decoder through an MLP to obtain the image quality score. QFD first accepts $f_{\text{up}}^{t,1}$, $f_{\text{up}}^{t,2}$, $f_{\text{up}}^{t,3}$, $f_{\text{up}}^{t,4}$ in $F_{\text{up}}^t$ as input, and upsamples all of them to a size of 64$\times$64. Next, a convolution layer and a squeeze-and-excite (SE) layer are used to unify the channel number to 512 for each feature map, and the four feature maps are concatenated into a single feature map with 2048 channels. This concatenated feature map is then processed through four convolution layers to gradually reduce the number of channels to 512, 128, 32, and 8. The QFD finally outputs an image quality feature map of size 64$\times$64$\times$8 as $F_{\text{quality}} = D(F_{\text{up}}^t)$. The $F_{\text{quality}}$ is flattened into a one-dimensional vector and passed through a regression network $R$, which consists of a three-layer MLP, to perform score regression and obtain the predicted value $y_p$. The process is as follows:
\begin{equation}
F_{\text{quality}} = D(F_{\text{up}}^t) = D(f_{\text{up}}^{t,1}, f_{\text{up}}^{t,2}, f_{\text{up}}^{t,3}, f_{\text{up}}^{t,4}),
\end{equation}
\begin{equation}
y_p = R(\mathrm{Flatten}(F_{\text{quality}})).
\end{equation}

\noindent\textbf{Model optimization.} Our model is trained in an end-to-end manner. The loss function consists of Mean Squared Error (MSE) loss $\mathcal{L}_{\text{mse}}$ and Margin loss $\mathcal{L}_{\text{mgn}}$, which are commonly used for learning image quality score regression and ranking (i.e., distinguishing the quality relationship within a batch) in IQA. Assuming the batch size is $n$, the GT image quality score is $y$, the predicted value is $y_p$, and the standard deviation of $y$ is $\sigma_y$, the loss functions are as follows:
\begin{equation}
\mathcal{L}_{\text{mse}} = \frac{1}{n} \| y - y_p \|_2^2,
\end{equation}
\begin{equation}
\mathcal{L}_{\text{mgn}} = \frac{2\sum_{i<j} \max \left( 0, -\mathrm{sign}(y_i - y_j) \cdot (y_{p_i} - y_{p_j}) + m \right)}{n(n-1)},
\end{equation}
where $m = \lambda \sigma_y$, $\lambda \in [0,1]$. Therefore, the overall loss function $\mathcal{L}_{\text{total}}$ can be defined as:
\begin{equation}
\label{l_teacher}
\mathcal{L}_{\text{total}}(y, y_p) = L_{\text{mse}}(y, y_p) + L_{\text{margin}}(y, y_p).
\end{equation}
This model is referred to as the ``teacher model'', and its loss function can also be written as $\mathcal{L}_{\text{teacher}}=\mathcal{L}_{\text{total}}(y, y_p)$.

\subsection{Knowledge distillation}
\label{student}
\textbf{Student model.} 
Unlike T2I models that require full image distribution modeling which requires a large network capcacity, our approach focuses on IQA, which inherently requires fewer parameters. To reduce the model's parameters and increase inference speed, we propose distilling the feature distribution of DP-IQA into a student model. We use a lightweight EfficientNet~\cite{EN} as the student model and adjust its output structure to align with that of the teacher model. By distillation, the student network only needs to learn the prior that corresponds to the image quality assessment. 


\noindent\textbf{Model optimization.} The student model takes the image as input and uses the output feature map $F_\text{quality}$ from the QFD as supervision to distill the image quality knowledge learned by the teacher model, we use the MSE shown in Equation (9) as the distillation loss $\mathcal{L}_{\text{disti}}$. Additionally, the student model is supervised by the GT image quality score $y$. Assuming the last feature map before the output layer of the student model is $F_\text{student}$, the predicted value of student model is $p_\text{ps}$, the loss function $\mathcal{L}_{\text{student}}$ for the student model can be defined as:
\begin{equation}
\label{l_student}
\mathcal{L}_{\text{student}} = \mathcal{L}_{\text{ disti}}(F_{\text{quality}}, F_{\text{student}}) + \mathcal{L}_{\text{total}}(y, y_\text{ps})
\end{equation}

\begin{table}[t]
  \caption{Learning rate decay at which epoch.}
  \centering
  \footnotesize
  \setlength{\tabcolsep}{2pt}
  \begin{tabular}{lllll}
    \toprule
    Model     & CLIVE     &  KonIQ &  LIVEFB &  SPAQ \\
    \midrule
    Teacher & - &5  &2 &-   \\
    Student &10, 25&5&4&6 \\
    \bottomrule
  \end{tabular}
  \label{lrd}
\end{table}

\begin{table}[t]
\caption{The values of the numeric variables defined in Sec.~\ref{sec:method}}
  \centering
  \footnotesize
  \setlength{\tabcolsep}{2pt}
  \begin{tabular}{llll}
    \toprule
    Variable     & Value& Explanation  \\
    \midrule

    $H$     & 512&  Height of the input image  \\
    $W$     & 512&  Width of the input image\\
    $l_s$     & 11&  The number of elements in \{scenes\} \\
    $l_d$     & 35&  The number of elements in \{distortion type\}\\
    $l_q$& 5&  The number of elements in \{quality level\}   \\
    $K$& 1925 & The total number of combinations of text templates       \\
    $d$& 768 &  Output dimension of the CLIP encoder        \\
    $\lambda$& 0.25& Coefficient used to control the margin \\
    \bottomrule
  \end{tabular}
  \label{pd}
\end{table}

\begin{table}[t]
\caption{Details of the text template we use in the experiment.}
  \centering
  \footnotesize
  \setlength{\tabcolsep}{2pt}
  \begin{tabular}{ll}
    \toprule
    Word types   & Details  \\
    \midrule
    \multirow{2}{*}{Scenes} & animal\, cityscape, human, indoor, landscape, \\
    &night, plant, still\_life, other\\
    \midrule 
    \multirow{9}{*}{Distortion type}  &jpeg2000 compression, jpeg compression, motion,\\
    &white noise, gaussian blur, fastfading, fnoise, lens, \\
    & diffusion, shifting, color quantization, desaturation \\
    & oversaturation, underexposure, overexposure, contrast,\\
    &white noise with color, impulse, multiplicative, jitter, \\
    &white noise with denoise, brighten, darken, pixelate, \\
    &shifting the mean, noneccentricity patch, quantization, \\
    &color blocking, sharpness, realistic blur, realistic noise,\\
    &realistic contrast change, other realistic, other\\
    \midrule 
    Quality level& bad, poor, fair, good, perfect\\
    \bottomrule
  \end{tabular}
    \label{text}
\end{table}

\section{Experiment}
\label{sec:exp}
\subsection{Datasets and evaluation metrics}
\textbf{Datasets.} IQA datasets primarily consist of distorted images paired with quality scores. We assess our DP-IQA using four in-the-wild IQA datasets:  CLIVE~\cite{CLIVE}, KonIQ~\cite{KonIQ}, LIVEFB (FLIVE)~\cite{2015} and SPAQ~\cite{SPAQ}, containing 1162, 10073, 11125, and 39810 authentically distorted (in-the-wild) images, respectively. Our research focuses on authentic distortion types, therefore artificially distorted datasets~\cite{PIPAL} such as LIVE~\cite{LIVE}, CSIQ~\cite{CSIQ} and KADID~\cite{KADID}, as well as artistic or stylized datasets, were not included in our study.


\noindent\textbf{Evaluation metrics.} Consistent with other works, we use Pearson's linear correlation coefficient (PLCC) and Spearman's rank-order correlation coefficient (SRCC) as performance evaluation metrics. PLCC measures the strength of a linear relationship between predicted and true values. Meanwhile, SRCC assesses the consistency of rank ordering between predicted and true values, focusing on monotonic relationships. 
Together, they provide a comprehensive evaluation, where higher values for both signify better performance.

\subsection{Implementation}
We implement our model using PyTorch and conduct training and testing on an A100 GPU. The version of stable diffusion is v1.5, while EfficientNet-B7 served as the backbone for the student model. We use Adam as the optimizer. The teacher model is trained with a batch size of at least 12, an initial learning rate of $10^{-5}$, for up to 15 epochs, while the student model is trained with a batch size of at least 24, an initial learning rate of $10^{-4}$ and for up to 30. Besides, the validation step for CLIVE is 50 while for other datasets is 250. Due to varying dataset scales, learning rate decay differ slightly across datasets, as detailed in Table~\ref{lrd}, and the scheduler is MultiStepLR, decay factor is 0.2. We also provide detailed values of the numeric variables defined in Sec.3 in Table~\ref{pd}. For data preprocessing, we resize in-the-wild images to 512$\times$512 pixels without patch splitting. We randomly split datasets into training and testing sets in 8:2, and repeat the splitting process five times for all datasets and report the median results. Besides, We present the specific settings of the template in Table~\ref{text}.

\begin{table*}
    \caption{Comparison of our proposed DP-IQA with SOTA BIQA algorithms on authentically distorted (in-the-wild) datasets. Bold entries indicate the top two results. '-' are not available publicly.}
  \centering
  \begin{tabular}{l|ll|ll|ll|ll}
    \toprule
    Dataset     & \multicolumn{2}{c}{CLIVE}     &  \multicolumn{2}{c}{KonIQ} &  \multicolumn{2}{c}{LIVEFB (FLIVE)} &  \multicolumn{2}{c}{SPAQ} \\
    \midrule 
    Metrics&PLCC&SRCC&PLCC&SRCC&PLCC&SRCC&PLCC&SRCC\\
    \midrule
    DIIVINE~\cite{DIIVINE} &0.591&0.588&0.558&0.546&0.187&0.092 &0.660&0.599\\
    BRISQUE~\cite{BRISQUE} &0.629&0.629&0.685&0.681&0.341&0.303 &0.817&0.809\\
    ILNIQE~\cite{ILNIQE}  &0.508&0.508&0.537&0.523&0.332&0.294 &0.712&0.713\\
    BIECON~\cite{206}  &0.613&0.613&0.654&0.651&0.428&0.407&-&-\\
    MEON~\cite{207}    &0.710&0.697&0.628&0.611&0.394&0.365&-&-\\
    WaDIQaM~\cite{WaDIQaM} &0.671&0.682&0.807&0.804&0.467&0.455&-&-\\
    DBCNN~\cite{DBCNN}   &0.869&0.851&0.884&0.875&0.551&0.545&0.915&0.911\\
    MetaIQA~\cite{MetaIQA} &0.802&0.835&0.856&0.887&0.507&0.540&-&-\\
    P2P-BM~\cite{2015}  &0.842&0.844&0.885&0.872&0.598&0.526&-&-\\
    HyperIQA~\cite{HyperIQA}&0.882&0.859&0.917&0.906&0.602&	0.544&0.915&0.911\\
    TIQA~\cite{TIQA}    &0.861&0.845&0.903&0.892&0.581&0.541&-&-\\
    MUSIQ~\cite{MUSIQ}&0.746&0.702&0.928&0.916&0.661&0.566&0.921&0.918\\
    TReS~\cite{TReS} &0.877&0.846&0.928&0.915&0.625&0.554&-&-\\
    DEIQT~\cite{DEIQT} &0.886&0.861&0.934&0.921&0.645&0.557&0.921&0.914\\
    CLIP-IQA~\cite{CLIPIQA} &0.832&0.805&0.909&0.895&-&-&0.866&0.864 \\
    ReIQA~\cite{ReIQA}   & 0.854&0.840&0.923&0.914&-&-&0.925&0.918\\
    SaTQA~\cite{SaTQA} & \textbf{0.903}&\textbf{0.877}&0.941&0.930&0.676&\textbf{0.582}&-&-\\
    {LIQE~\cite{LIQE}}  & {-}&{-}&{0.912}&{0.928}&{-}&{-}&{0.919}&{0.922}\\
    {Q-Align~\cite{QAlign}}  & {-}&{-}&{0.941}&{\textbf{0.940}}&{-}&{-}&{\textbf{0.933}}&{\textbf{0.930}}\\
    LoDa~\cite{LoDa}  & 0.899&0.876&\textbf{0.944}&0.932&\textbf{0.679}&0.578&\textbf{0.928}&\textbf{0.925}\\
    \midrule
    Ours (student) & 0.902& 0.875&\textbf{0.944}&0.926&0.671&0.567	&0.923&0.920\\
    Ours (teacher)  &\textbf{0.913}&\textbf{0.893}	&\textbf{0.951}&\textbf{0.942}	&\textbf{0.683}&\textbf{0.579}&0.926&0.923\\
    \bottomrule
  \end{tabular}
    \label{cp1}
    \vspace{-0.1 in}
\end{table*}
\begin{table}
    \caption{{Comparison of SRCC on cross datasets setting, \textit{i.e.}, we test and report the performance of models on unseen datasets. Bold entries indicate the top two results.'-' are not available publicly.}}
  \centering
  \begin{tabular}{l|ll|l|l}
    \toprule
    Training on &  \multicolumn{2}{c|}{LIVEFB} & CLIVE     &  KonIQ \\
    \midrule 
    Testing on&  KonIQ & CLIVE&  KonIQ & CLIVE\\
    \midrule
    DBCNN&0.716&0.724&0.754&0.755\\
    P2P-BM&0.755&0.738&0.740&0.770\\
    HyperIQA&0.758&0.735&0.772&0.785\\
    TReS&0.713&0.740&0.733&0.786\\
    SaTQA&-&-&\textbf{0.788}&0.791\\
    {Q-Align}&{-}&{-}&{-}&{\textbf{0.853}}\\
    LoDa&0.763&\textbf{0.805}&0.745&0.811\\
    \midrule
    Ours (student) & \textbf{0.767}	&0.758&\textbf{0.781} &0.830 \\
    Ours (teacher)  &\textbf{0.771}	&\textbf{0.770}	&0.766	&\textbf{0.833}\\
    \bottomrule
  \end{tabular}
    \label{cp2}
        \vspace{-0.1 in}
\end{table}
\begin{table}
    \caption{{The results of regression on the output features of different pre-trained backbones with frozen parameters to evaluate image quality.}}
  \centering
  \footnotesize
  \setlength{\tabcolsep}{4pt}
  \begin{tabular}{l|c|cc|cc}
    \toprule
    \multirow{2}{*}{Backbone} &Learning& \multicolumn{2}{c}{CLIVE}& \multicolumn{2}{c}{KonIQ} \\
    \cmidrule{3-6} 
     &Paradigm& PLCC & SRCC& PLCC & SRCC \\
    \midrule
    CLIP (b/16)~\cite{CLIP}&& 0.758& 0.765&0.828&0.800\\
    MAE (b/16)~\cite{MAE}&Self-supervised&0.640 & 0.618&0.720&0.689\\
    DINOv2 (b/14)~\cite{DINOv2}&&0.626 & 0.600&0.701&0.665\\
    \midrule
    ViT (b/16)~\cite{ViT}&&0.524&0.502&0.682&0.646\\
    ResNet-50~\cite{RN50}&Supervised&0.770&0.767&0.837&0.821\\
    Stable Diffusion~\cite{SD}&&\textbf{0.869} & \textbf{0.817}&\textbf{0.929}&\textbf{0.908}\\
    \bottomrule
  \end{tabular}
    \label{cp3}
        \vspace{-0.1 in}
\end{table}

\begin{table}
      \caption{Ablation analysis of text prompt (TP), constant conditional embedding (CCE), text adapter (TA) and image adapter (IA) in teacher model. Bold entries indicate the best results.}
  \centering
  \setlength{\tabcolsep}{1pt}
  \begin{tabular}{l|ll|ll|ll|ll|ll}
    \toprule
     \multirow{2}{*}{Dataset} &  \multicolumn{2}{c|}{Full} & \multicolumn{2}{c|}{w/o TP}  &\multicolumn{2}{c|}{w/o CCE} & \multicolumn{2}{c|}{w/o TA} &\multicolumn{2}{c}{{w/o IA}}\\
    \cmidrule(lr){2-11} 
     &  PLCC & SRCC&  PLCC & SRCC&  PLCC & SRCC&  PLCC & SRCC&  PLCC & SRCC\\
    \midrule
    CLIVE&\textbf{0.913}&\textbf{0.893}&0.867&0.871&0.898&0.878&0.907&0.881&0.904&0.875\\
    KonIQ&\textbf{0.951}&\textbf{0.942}&0.929&0.931&0.937&0.928&0.941&0.940&0.946&0.932\\
    \bottomrule
  \end{tabular}
     \label{ab1}
         \vspace{-0.1 in}
\end{table}

\subsection{Comparison against other methods}

\textbf{Overall comparison.} We compare our method with 18 SOTA baselines\footnote{Preprints and works w/o released code are not included in the comparison. 
{Besides, we do not include comparisons with works that use customized experimental settings, such as joint training on multiple datasets or other conditions.}}
Table~\ref{cp1} compares the performance of our method with others across four widely recognized in-the-wild datasets. Results for DEIQT~\cite{DEIQT} are based on our reproduction, while those for other methods are taken from the original papers of LoDa~\cite{LoDa}, TReS~\cite{TReS}, and SaTQA~\cite{SaTQA}. The experimental findings demonstrate that our proposed method achieves SOTA performance on the CLIVE (LIVEC), KonIQ, and LIVEFB (FLIVE) datasets, underscoring its effectiveness across diverse real-world scenarios. Additionally, on the SPAQ dataset, our method delivers highly competitive results that are nearly on par with the best-performing approach. Furthermore, the student model exhibits only a marginal decline in performance across these datasets. This outcome aligns with expectations and highlights the success of the distillation process in effectively transferring knowledge.


\noindent\textbf{Generalization ability.} The practical value of a model is closely tied to its generalization capability. In Table~\ref{cp2}, we evaluate our model's generalization through cross-dataset zero-shot performance on three in-the-wild datasets. Training is done on one dataset, and testing on unseen datasets. We compare our method with SOTA baselines, and the results show superior generalization in most cases. Additionally, the student model performs similarly to the teacher model but with far fewer parameters, demonstrating its practical value. For larger datasets (KonIQ, LIVEFB, and SPAQ), the student model's performance closely matches the teacher model, indicating effective knowledge distillation. However, for smaller dataset (CLIVE), the student model outperforms the teacher model. This may be because the teacher model is too large and poses a higher risk of overfitting on very small datasets, whereas the use of the student model significantly ameliorated this issue, improving generalizability.

\noindent{\noindent\textbf{Effectiveness of different priors.} In Table~\ref{cp3}, we use linear probing to assess the prior effectiveness of various pretrained backbones for IQA. We freeze the backbone parameters and train only a linear regressor on features from supervised models like ResNet-50~\cite{RN50} and ViT~\cite{ViT}, and self-supervised ones such as CLIP~\cite{CLIP}, MAE~\cite{MAE}, and DINOv2~\cite{DINOv2}. 
Results show that diffusion-based representations are more effective for IQA. We hypothesize that this is because the learning strategy of the T2I diffusion model allows its prior to be viewed as the learned gradient of data density, which contains rich low-level information. Additionally, with natural language supervision, the model also incorporates significant high-level semantic knowledge. Such combined prior enables the T2I diffusion model to achieve better performance in IQA tasks.}

\begin{table}[t]
  \caption{Ablation analysis of the settings of timestep for teacher model. Bold entries indicate the best results.}
  \centering
  \setlength{\tabcolsep}{1pt}
  \begin{tabular}{l|ll|ll|ll|ll|ll}
    \toprule
     \multirow{3}{*}{Dataset} &  \multicolumn{9}{c}{Timestep} \\ \cmidrule{2-11}
     &  \multicolumn{2}{c|}{1} &  \multicolumn{2}{c|}{5} &\multicolumn{2}{c|}{10} & \multicolumn{2}{c|}{20}& \multicolumn{2}{c}{50} \\ \cmidrule{2-11}
     &  PLCC & SRCC&  PLCC & SRCC&  PLCC & SRCC&  PLCC & SRCC&  PLCC & SRCC\\ 
    \midrule
    CLIVE&\textbf{0.913}&\textbf{0.893}&\textbf{0.913}&\textbf{0.893}&0.912&0.879&\textbf{0.913}&0.879&0.907&0.871\\
    KonIQ&\textbf{0.951}&\textbf{0.942}&0.947&0.939&0.945&0.936&0.946&0.936&0.942&0.931\\
    \bottomrule
  \end{tabular}
      \label{ab2}
\end{table}

\begin{table}[t]
    \caption{{Ablation analysis of the multi-level feature extraction strategy, where the timestep $t=1$. Bold entries indicate the best results.}}
  \centering
    \setlength{\tabcolsep}{1pt}
  \begin{tabular}{l|ll|ll|ll|ll|ll}
    \toprule
     \multirow{2}{*}{Dataset} &  \multicolumn{2}{c|}{Full} & \multicolumn{2}{c|}{only $f_{\text{up}}^{t,1}$}  & \multicolumn{2}{c|}{only $f_{\text{up}}^{t,2}$} &\multicolumn{2}{c|}{only $f_{\text{up}}^{t,3}$}&\multicolumn{2}{c}{only $f_{\text{up}}^{t,4}$}\\
    \cmidrule(lr){2-11} 
     &  PLCC & SRCC&  PLCC & SRCC&  PLCC & SRCC&  PLCC & SRCC&  PLCC & SRCC\\
    \midrule
    CLIVE&\textbf{0.913}&\textbf{0.893}&0.867&0.812&0.874&0.845&0.879&0.841&0.869&0.817\\
    KonIQ&\textbf{0.951}&\textbf{0.942}&0.923&0.903&0.937&0.921&0.941&0.927&0.929&0.908\\
    \bottomrule
  \end{tabular}
\label{mfe}
\end{table}

\begin{table}[t]
    \caption{Ablation analysis of feature extracted from each layer, where the timestep $t=1$. Bold entries indicate the best results.}
  \centering
  \setlength{\tabcolsep}{1pt}
  \begin{tabular}{l|ll|ll|ll|ll|ll}
    \toprule
     \multirow{2}{*}{Dataset} &  \multicolumn{2}{c|}{Full} & \multicolumn{2}{c|}{w/o $f_{\text{up}}^{t,1}$}  & \multicolumn{2}{c|}{w/o $f_{\text{up}}^{t,2}$} &\multicolumn{2}{c|}{w/o $f_{\text{up}}^{t,3}$}&\multicolumn{2}{c}{w/o $f_{\text{up}}^{t,4}$}\\
    \cmidrule(lr){2-11} 
     &  PLCC & SRCC&  PLCC & SRCC&  PLCC & SRCC&  PLCC & SRCC&  PLCC & SRCC\\
    \midrule
    CLIVE&\textbf{0.913}&\textbf{0.893}&0.909&0.891&0.904&0.875&0.897&0.874&0.910&\textbf{0.893}\\
    KonIQ&\textbf{0.951}&\textbf{0.942}&\textbf{0.951}&0.939&0.947&0.941&0.945&0.936&0.949&\textbf{0.942}\\
    \bottomrule
  \end{tabular}
\label{mfe2}
\end{table}


\begin{table}[t]
    \caption{Ablation analysis of the distillation loss $\mathcal{L}_{\text{disti}}$. Using the distillation can significantly improve the performance than training using $L_{\text{teacher}}$ under the same lightweight backbone. Bold entries indicate the best results.}
  \centering
  \begin{tabular}{l|cc|cc}
    \toprule
     \multirow{2}{*}{Dataset} &  \multicolumn{2}{c|}{Distilled student} & \multicolumn{2}{c}{w/o distillation loss} \\
    \cmidrule{2-5} 
     &  PLCC & SRCC&  PLCC & SRCC\\
    \midrule
    CLIVE&\textbf{0.902}&\textbf{0.875}&0.717&0.715\\
    KonIQ&\textbf{0.944}&\textbf{0.926}&0.881&0.841\\
    \bottomrule
  \end{tabular}
     \label{ab3}
\end{table}

\subsection{Ablation}

\noindent\textbf{Text prompt and adapters.} As shown in Table~\ref{ab1}, we explore the impact of text prompt and constant conditional embedding strategy, and ``w/o CCE'' means using the description in the template that best matches the current image content as input, rather than using all templates. Additionally, we also conduct ablation studies on the text and image adapters. When there is no text prompt (w/o TP), the text adapter was not activated by default. The results indicate that the text prompt, constant conditional embedding strategy, image adapter, and text adapter play positive roles in overall performance. 

\noindent\textbf{Timesteps.} We observe the impact of different timestep settings on model performance. As shown in Table~\ref{ab2}, using smaller timesteps is generally more advantageous. Therefore, we set $t=1$ in our other experiments. However, considering that the model's prior knowledge is embedded in its learned parameters, the choice of time steps does not have a decisive impact in the case of full parameter fine-tuning.

\noindent\textbf{Multi-level features.} {As shown in Table~\ref{mfe}, we conduct ablation analysis on the multi-level feature extraction strategy, where we use only one layer of features for image quality assessment. The experimental results show that using features from a single layer leads to a performance drop, which highlights the importance of multi-scale features. Moreover, as shown in Table~\ref{mfe2}, we find that each level of features positively impact the results, with $f_{\text{up}}^{t,2}$ and $f_{\text{up}}^{t,3}$ being potentially more important. However, although the impact of features from different layers varies, discarding any layer leads to performance degradation, indicating that the contributions of features from different layers are partially orthogonal and indispensable.}


\begin{figure*}[t]
  \centering
  \includegraphics[width=\linewidth]{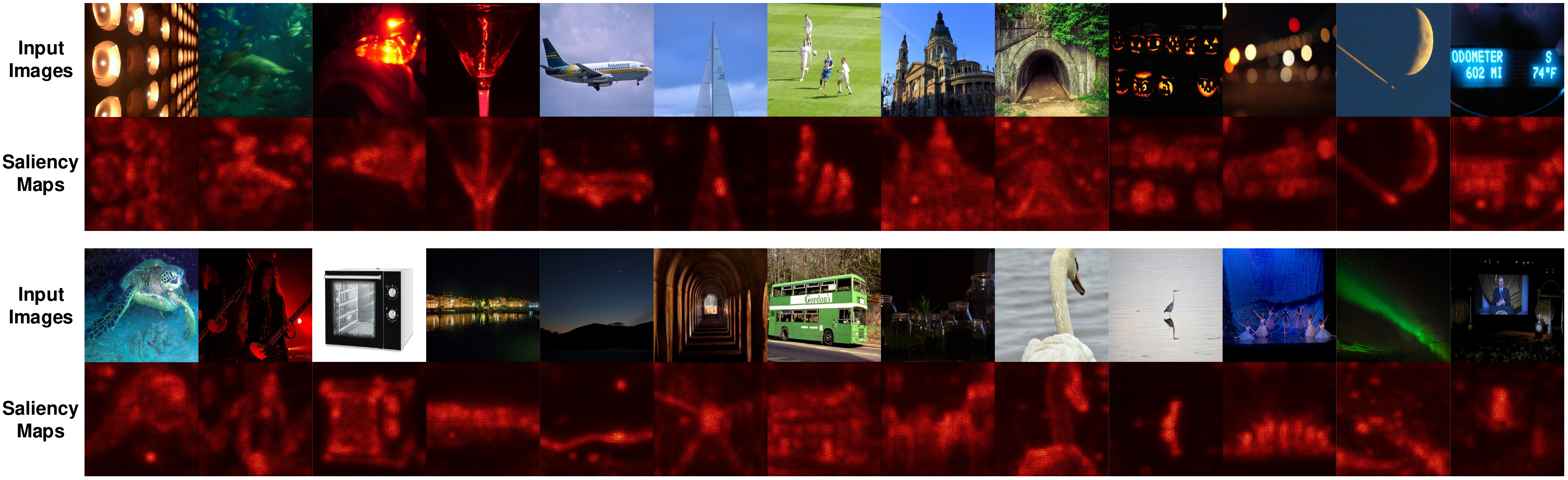}
  \caption{Saliency maps generated by DP-IQA on in-the-wild images from the KonIQ-10k dataset. The model effectively prioritizes complex structures and semantically significant regions, aligning with human visual perception. Additionally, the model does not exhibit significant overfitting to noise, demonstrating strong noise resistance.}
  \label{smap}
\end{figure*}

\begin{figure}[t]
  \centering
  \includegraphics[width=0.85\linewidth]{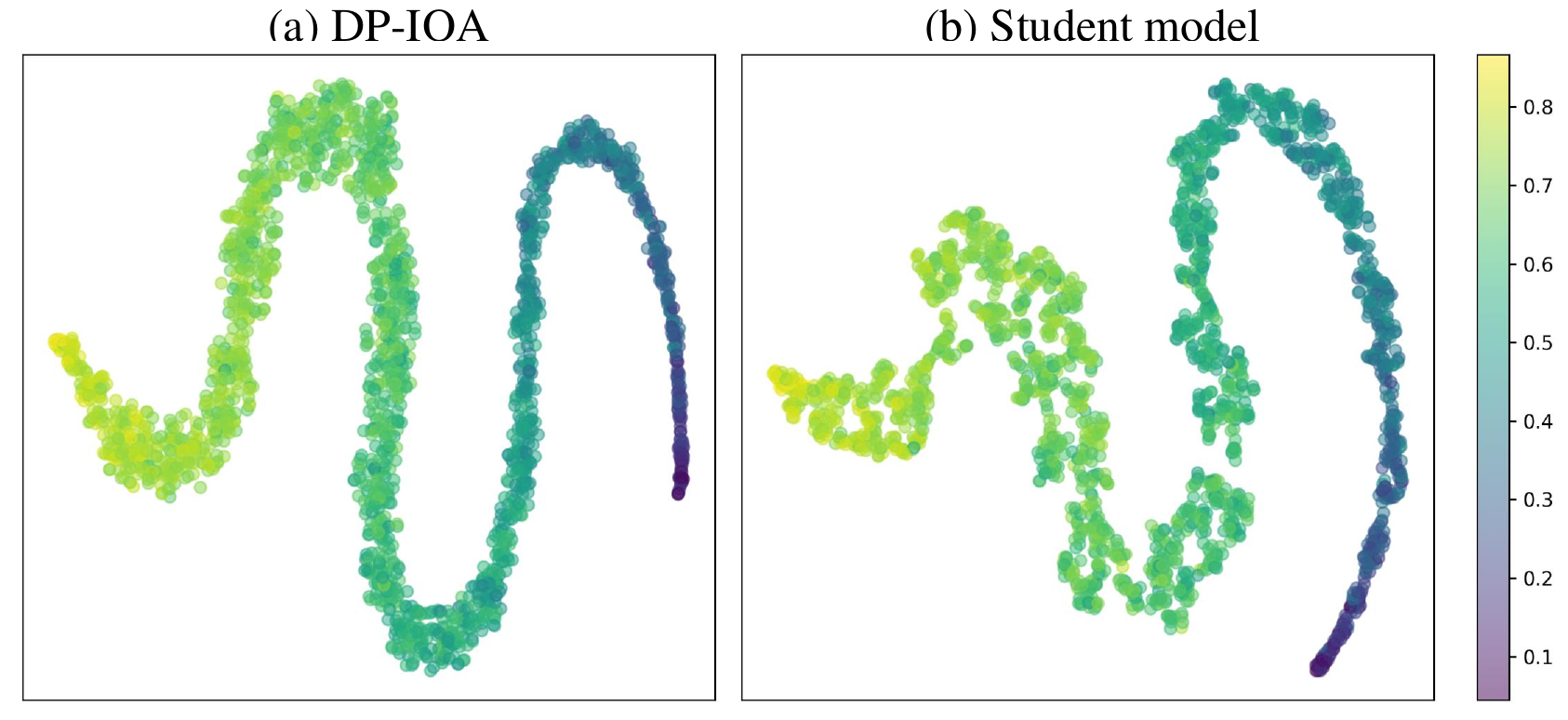}
  \caption{t-SNE visualization of the feature embeddings from the last hidden layer of DP-IQA and its student models. The input images (points in the subplots) are from the test set of Koniq-10k, with the color gradient representing their ground truth scores (from 0 to 1, with higher values indicating better image quality).}
  \label{tsne}
\end{figure}

\noindent\textbf{Distillation.} As shown in Table~\ref{ab3}, we conduct ablation analysis on the distillation. Experimental results indicate that distillation effectively enhances the performance of the student model. The results show the effectiveness of distilling enhanced priors from DP-IQA compared with training from scratch, which also demonstrates that the outstanding performance of the student model does not solely rely on its own architecture and pre-trained weights. {A comparison of the number of parameters and inference speed is shown in Table~\ref{tab:tp}, where student model achieves similar performance with $\sim3\times$ speed up and $\sim14\times$ size reduction.}

\noindent{\noindent\textbf{Training strategies and pre-trained priors.} As shown in Table~\ref{ab4}, we conducted an ablation study to assess the effects of training strategies and pretrained priors. For training strategies, we examined freezing the pre-trained U-Net and training QFD and adapters, as well as applying LoRA~\cite{LoRA}. Although these approaches did not surpass SOTA methods, they achieved reasonable performance, indicating the value of the pre-trained SD prior for IQA. However, compared to full-parameter fine-tuning, training from scratch (only training QFD and adapters or using LoRA) required significantly more steps to converge, while offering limited computational savings and encountering notable performance bottlenecks. Moreover, the results highlight the importance of leveraging pretrained SD weights, as models without them struggled to extract meaningful features from limited IQA data, further demonstrating the effectiveness of pre-trained priors.}

\section{Discussion}
\subsection{Visual saliency}
{The cross-attention map in Figure~\ref{moti} shows that the T2I diffusion model can attend to both high-level semantics and low-level distortions through the text prompt, indicating its reliable prior knowledge.} However, it does not explain how the model evaluates perceptual quality or assigns scores. Understanding this process is crucial for assessing whether the fine-tuned model aligns with human visual perception and its sensitivity to noise in small-scale BIQA datasets, which is vital for generalization. {To better illustrate the model's behavior, we visualize saliency maps to show which regions influence its decision-making. These saliency maps highlight pixel-level regions that affect the model's output, offering different insights than the cross-attention maps.}

As shown in Figure~\ref{smap}, the model tends to focus on complex structures, semantically important objects, and areas with significant color or brightness variation. {This aligns with human visual perception, where attention is drawn to high-contrast or semantically meaningful regions.} The model's ability to focus on these key features suggests it has learned important cues for IQA tasks. Additionally, we observed no significant overfitting to noise, highlighting the model's strong noise resistance and its excellent cross-dataset generalization, which enhances its adaptability and reliability in diverse scenarios.

\begin{table}[t]
  \caption{
  {The average time spent per
image on our hardware platform and the number of parameters between our teacher and student model. Bold entries indicate the best results.}}
  \centering
  \begin{tabular}{l|cc}
    \toprule
     Model  & Time (s/image) & Params\\
    \midrule
     DP-IQA (teacher) &0.023&1.19B\\
     Distilled student &\textbf{0.006}&\textbf{81.01M} \\
    \bottomrule
  \end{tabular}
\label{tab:tp}
\vspace{-0.1 in}
\end{table}

\begin{table}[t]
    \caption{{Ablation study on the impact of pretrained priors and training strategies on DP-IQA's performance.}}
  \centering
    \scriptsize
  \setlength{\tabcolsep}{3pt}
  \begin{tabular}{l|c|cc|cc}
    \toprule
    \multirow{2}{*}{Trainable Params} &\multirow{2}{*}{Pretrained U-Net}&\multicolumn{2}{c}{CLIVE}&\multicolumn{2}{c}{KonIQ}\\
    \cmidrule{3-6} 
     && PLCC & SRCC & PLCC & SRCC   \\
    \midrule
    QFD&\checkmark&0.884 &0.849&0.929 &0.908 \\
    Adapters+QFD&\checkmark&0.890&0.829&0.935&0.917\\
    LoRA+QFD&\checkmark&0.903&0.858 &0.943 &0.928 \\
    Adapters+LoRA+QFD&\checkmark&0.907&0.872&0.946 &0.931\\
    Full&\checkmark& \textbf{0.913}& \textbf{0.893}&\textbf{0.951}&\textbf{0.942}\\
    Full&$\times$&0.551 & 0.569 &0.786 &0.784\\
    \bottomrule
  \end{tabular}
    \label{ab4}
\end{table}

\subsection{t-SNE Visualization}
As illustrated in Figure~\ref{tsne}, t-Distributed Stochastic Neighbor Embedding (t-SNE) is employed to visualize the feature embeddings from the final hidden layers of DP-IQA and its corresponding student models, using the complete set of test images from the Koniq-10k dataset as input. A gradient of colors is used to represent the ground truth (GT) scores, facilitating the exploration of the relationship between the learned features and image quality. As a non-linear dimensionality reduction technique, t-SNE projects high-dimensional features into a two-dimensional space, preserving the local structural relationships among data points. If images with similar GT scores are clustered in proximity, it suggests that the model has effectively learned features that are strongly correlated with image quality. The resulting scatter plot exhibits a well-defined, continuous color gradient, which signifies a strong correspondence between the GT scores and the spatial arrangement of the feature embeddings, thereby demonstrating the model's capacity to discern subtle variations in image quality.

\subsection{Limitations}
As shown in Table~\ref{tab:tp}, we have distilled the knowledge of DP-IQA into a relatively lightweight model that is both easy to deploy and fast. However, further reducing its size while maintaining prediction accuracy and generalization remains a significant challenge. 
Further investigation is needed to develop more effective approaches for distilling image-quality-related prior knowledge from teacher models into lightweight student models. {Although DP-IQA is designed for in-the-wild scenarios, we also evaluate its performance in synthetic distortion scenarios. We observed that the DP-IQA teacher model may overfit on small-scale synthetic datasets. As synthetic distortion is not the focus of this paper, further details are provided in the supplementary materials.}



\section{Conclusion}

In this paper, we propose a novel BIQA method based on large-scale pre-trained diffusion priors for in-the-wild images, named DP-IQA. It leverages pre-trained SD as the backbone, extracting multi-level features from the denoising U-Net during the upsampling process at a specific timestep and decoding them to estimate image quality, without requiring a diffusion process. To alleviate the computational burden of diffusion models in practical applications, we distill the knowledge from DP-IQA into a smaller EfficientNet-based model. Experimental results show that DP-IQA achieves SOTA on various in-the-wild datasets and demonstrates the best generalization capabilities. We believe our exploration can provide a new technical direction for future works and inspire future efforts to more effectively leverage diffusion priors for better assessment of image perceptual quality.

\bibliographystyle{IEEEtran}
\bibliography{references}

\end{document}